\documentclass{article}
\usepackage{acra}
\usepackage{graphicx}
\usepackage{multirow}
\usepackage{adjustbox}
\usepackage{amsmath}
\usepackage{url}
\usepackage{hyperref}
\usepackage{array}
\usepackage{amsmath,amssymb,amsfonts}
\usepackage{algorithmic}
\usepackage{graphicx}
\usepackage{textcomp}
\usepackage{xcolor}
\usepackage{caption}
\usepackage{subcaption}
\usepackage{ragged2e}
\title{Detection of Spider Mites on Labrador Beans through Machine Learning Approaches Using Custom Datasets}
\author{Violet Liu$^{1*}$, Jason Chen$^{1*}$, Ans Qureshi$^{1*}$, Mahla Nejati$^{1**}$\\Centre for Automation and Robotic Engineering Science, \\The University of Auckland, New Zealand $^1$\\ 
hliu665, jche774, aqur476@aucklanduni.ac.nz$^{*}$, m.nejati@auckland.ac.nz$^{**}$}

\begin{document}
% begin{pagenumber} for keeping track of page numbers, delete before submission
% \pagenumbering{arabic} 
% \pagestyle{plain}
% end{pagenumber}

\maketitle

\begin{abstract}
Amidst growing food production demands, early plant disease detection is essential to safeguard crops; this study proposes a visual machine learning approach for plant disease detection, harnessing RGB and NIR data collected in real-world conditions through a JAI FS-1600D-10GE camera to build an RGBN dataset. A two-stage early plant disease detection model with YOLOv8 and a sequential CNN was used to train on a dataset with partial labels, which showed a 3.6\% increase in mAP compared to a single-stage end-to-end segmentation model. The sequential CNN model achieved 90.62\% validation accuracy utilising RGBN data. An average of 6.25\% validation accuracy increase is found using RGBN in classification compared to RGB using ResNet15 and the sequential CNN models. Further research and dataset improvements are needed to meet food production demands.
\end{abstract}

\section{Introduction}

As the global population grows, the demand for food and crops also increases. It is estimated that a 70\% increase in food would be needed to meet our population's needs by 2050 \cite{fang2015current}. In 2021, New Zealand's horticulture industry made 6.68 billion dollars in exports, which makes 11.1\% of the country's total exports \cite{FreshFacts}. Despite efforts to enhance yield through automation in pollination \cite{Barnett2017,Nejati2019}, pruning \cite{Williams2023} and harvesting \cite{Nejati2019a}, pests and diseases still result in 20-40\% harvest losses \cite{thakur2022trends}. Furthermore, it is estimated that billions of dollars are lost yearly just in New Zealand due to the deterioration of plants caused by disease \cite{thakur2022trends}. The current leading industry method in disease detection is manual auditing, a time-consuming and labour-intensive process requiring trained professionals; this results in high labour costs for farmers. To alleviate the stress of disease detection, researchers are exploring different plant disease detection techniques, including using various sensors, data collection methods, and machine learning algorithms. Bioforce, a New Zealand natural pest control company, funds this study. The data used in this study is collected within their greenhouses. This study focuses on the early detection of spider mites in labrador beans using RGB and Near-infrared (NIR) data through machine learning algorithms. A two-stage model is proposed, pipelining a You Only Look Once version 8 (YOLOv8) segmentation model to a simple 6-layer sequential Convolutional Neural Network (CNN). The study delves into challenges with current plant disease detection datasets, data processing techniques, segmentation and classification methods. The primary contributions are:
\begin{itemize}
    \item Collecting a customised RGB and Near Infrared (RGBN) dataset for spider mites on labrador beans
    \item An examination of how RGBN datasets impact early disease detection in contrast to RGB datasets and the potential of partial transfer learning in RGBN input layer
    \item Analysis upon the effects of small datasets on high complexity CNN such as VGG16 and ResNet50
    \item Investigation on the effectiveness of a two-stage model on datasets with missing labels
\end{itemize}

The subsequent sections encompass \textbf{Related Works} on RGBN-based machine learning, dataset biases, segmentation and classification models. The \textbf{Methodology} section delves into design choices, providing rationales. \textbf{Results} presents quantitative data, the \textbf{Discussion} involves in-depth data analysis, and the \textbf{Conclusion} summarises findings. Additionally, it touches upon future prospects and acknowledgements.

\section{Related Works}
The study of plant disease detection through image processing and machine learning is a prominent and actively researched topic. Plant disease detection has many facets, including different data channels, usage of datasets, and state-of-the-art machine learning methods.
\subsection{Utilising RGB and NIR Channels in Data}
NIR data has been used to classify tomato leaves through colour image analysis and machine learning \cite{Nieuwenhuizen2020}. In this study a camera was placed in a greenhouse to capture images. The study utilised near-infrared bands; however, it found that Linear Discriminant Analysis (LDA) did not have the spatial resolution to identify early spider mite damage within these bands. After three months of spider mite damage, a ResNet18 model successfully attained a 90\% validation accuracy in distinguishing between "Healthy" and "Damaged" leaves through exclusively RGB data. Nevertheless, this study did not incorporate near-infrared data into the ResNet18 model due to hardware limitations. 

RGB and VIS-NIR spectral imaging has been used to detect "Apple Scab" disease in an apple orchid \cite{rous}. 8 channel based data was collected which consisted of, RGB and 5 different NIR frequencies. The study achieved a Mean Average Precision (mAP) of 0.73 by using only the RGB images with a YOLOv5 classification network; the reason why RGB was used is that it was challenging to balance the light conditions for all 8 channels in a multispectral camera in an outdoor environment.

A combination of RGB and NIR data channels, or RGBN channels, was found to improve the classification of kiwifruit detection by 1.5\% to 2.3\% \cite{Liu2020}; this was done by comparing the average precision of a trained VGG16 model by using different inputs such as RGB, NIR or a combination of the two through image or feature fusion. Although this study is not on plant disease detection and only detects whether a kiwifruit is present, it still reveals the potential benefits of utilising NIR data in classification and segmentation. 

Among these studies, none have addressed the possibility of classifying leaves based on stress-related abnormalities distinct from disease symptoms. Additionally, none have effectively incorporated RGBN data for early plant disease detection. 

\subsection{Dataset exploration}
Beginning with a discussion of publicly available datasets, it is found that the popularly used "Plant village" dataset is biased in its labelled classifications due to consistent background colors in the images \cite{Bhandari2023}. High validation accuracies of 90\% - 98\% and above were achieved overall within the studies that utilise it\cite{Noyan2022}. However, it was found that the background colours in the dataset create bias. Despite the dataset containing 38 categories, a validation accuracy of 49\%  was achieved using the random forest model; this is problematic as it proves a correlation between the background and the classes. Similar tests were conducted on the "MNIST" dataset as a benchmark, achieving a validation accuracy of 11.7\%; since the "MNIST" dataset only has 10 classes, 11.7\% is an appropriate value for random guesses, showing no bias. Similarly, many papers utilising the "NaCRRI" Ugandan public dataset have consistently achieved a validation accuracy of above 98\%, which is considered suspicious as it seems similar to the "Plant village" bias \cite{Devi2023,Elfatimi2022,Abed2021,Singh2023}. The "NaCRRI" Ugandan public dataset tends to have variations in data collection; for example, the leaf would be plucked, and at other times, it would be on the plant. It is plausible that these variations strengthen the dataset and enable the algorithms to achieve a higher validation accuracy. However, whether the dataset has an inherent bias remains unclear, as it is relatively new and unexplored. \cite{Elfatimi2022}.

Whilst it is problematic that these public datasets are potentially biased, collecting and labelling this data is incredibly time-consuming and labour-intensive. Due to the expensive nature of creating customised datasets for each study, utilising a public dataset to compare and test the functionalities of proposed models should still be encouraged.

\subsection{State-of-the-art Segmentation Models}
YOLOv8 is still being heavily applied in agriculture for detecting plant diseases. It showed robustness by achieving an F1-score of 99.2\% for segmentation on "PlantVillage" and "PlantDoc" datasets that consisted of 11700 RGB images with 5 plant species and 8 disease classes \cite{Qadri2023}. Recognising that some datasets do not faithfully represent the real world is imperative, as "PlantVillage" data is from controlled laboratory conditions.

\subsection{State-of-the-art Classification Models}
VGG16 model remains one of the preferred choices for plant disease detection, achieving a validation accuracy rate of 95.2\%  on the "Plantvillage" dataset. \cite{Alatawi2022} covered 19 distinct plant disease classes, spanning tomatoes, grapes, apples, and corn crops. \cite{Shah2022} compared the performance of Inception V3, VGG 16, VGG 19, CNN, and ResNet50 in early rice disease detection. In this study, ResNet50 achieved the highest validation accuracy of 99.75\% , with VGG16 close behind and a validation accuracy of 98.47\% . 

In another study \cite{saeed2021multi}, rice leaf datasets obtained from "Kaggle" were categorized into major and minor disease groups to understand the diversity of diseases better. These datasets were then used to train a variant of the ResNet152 model, resulting in an accuracy of 99.1\% for major disease detection and 82.20\% for minor disease detection. 

A notable limitation in these studies is their reliance on publicly available datasets, specifically "PlantVillage" and "Kaggle"; this limitation poses a challenge when applying their models in real-world scenarios because the "PlantVillage" and "Kaggle" datasets consist of photographs captured under controlled laboratory conditions rather than actual field conditions.

In another study, ResNet20, which is a relatively complex CNN model, exhibited inferior performance compared to a basic sequential CNN model when the dataset consisted of fewer than 640 images. This was found through the creation of three CNN models with different trainable parameter capacities, including one referred to as "CNN-Low Capacity", with the lowest number of trainable parameters, "CNN-Medium Capacity," "CNN-High Capacity," and "ResNet-20," boasting the highest number of trainable parameters\cite{Brigato2021}. The models were then tested on the sCIFAR10, sFMNIST, and sSVHN datasets. Interestingly, the only state-of-the-art model tested was ResNet, potentially due to other models being very complex. However, since only ResNet is tested as an example for "High number of trainable parameters", it could be possible that the inability to perform well with a dataset with less than 640 images is exclusive to ResNet.

\section{Methodology}
\begin{figure*}[t]
    \centering
    \begin{subfigure}{0.32\textwidth}
        \centering
        \includegraphics[width=\linewidth]{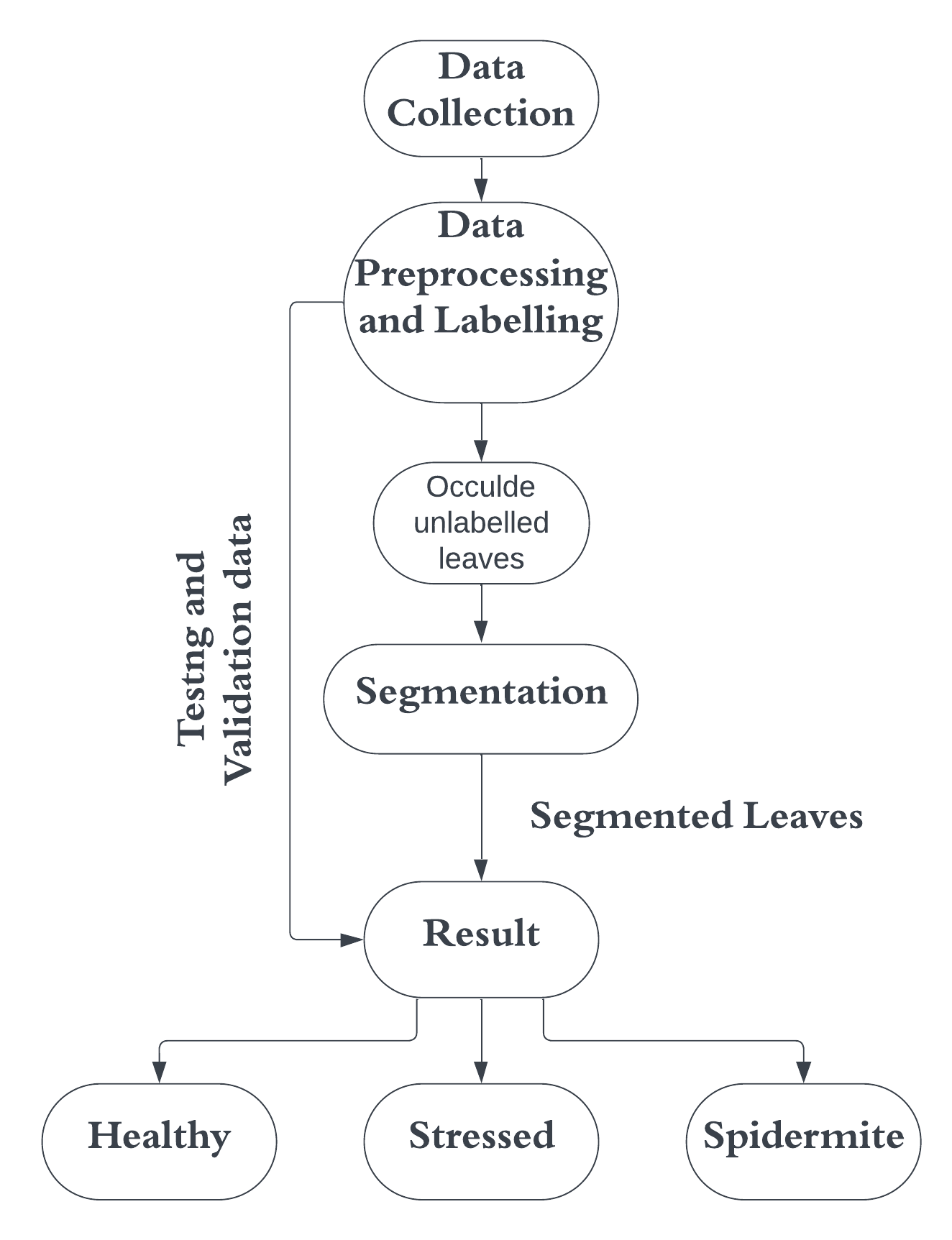}
        \caption{Single-Stage Model}
    \end{subfigure}
    \begin{subfigure}{0.32\textwidth}
        \centering
        \includegraphics[width=\linewidth]{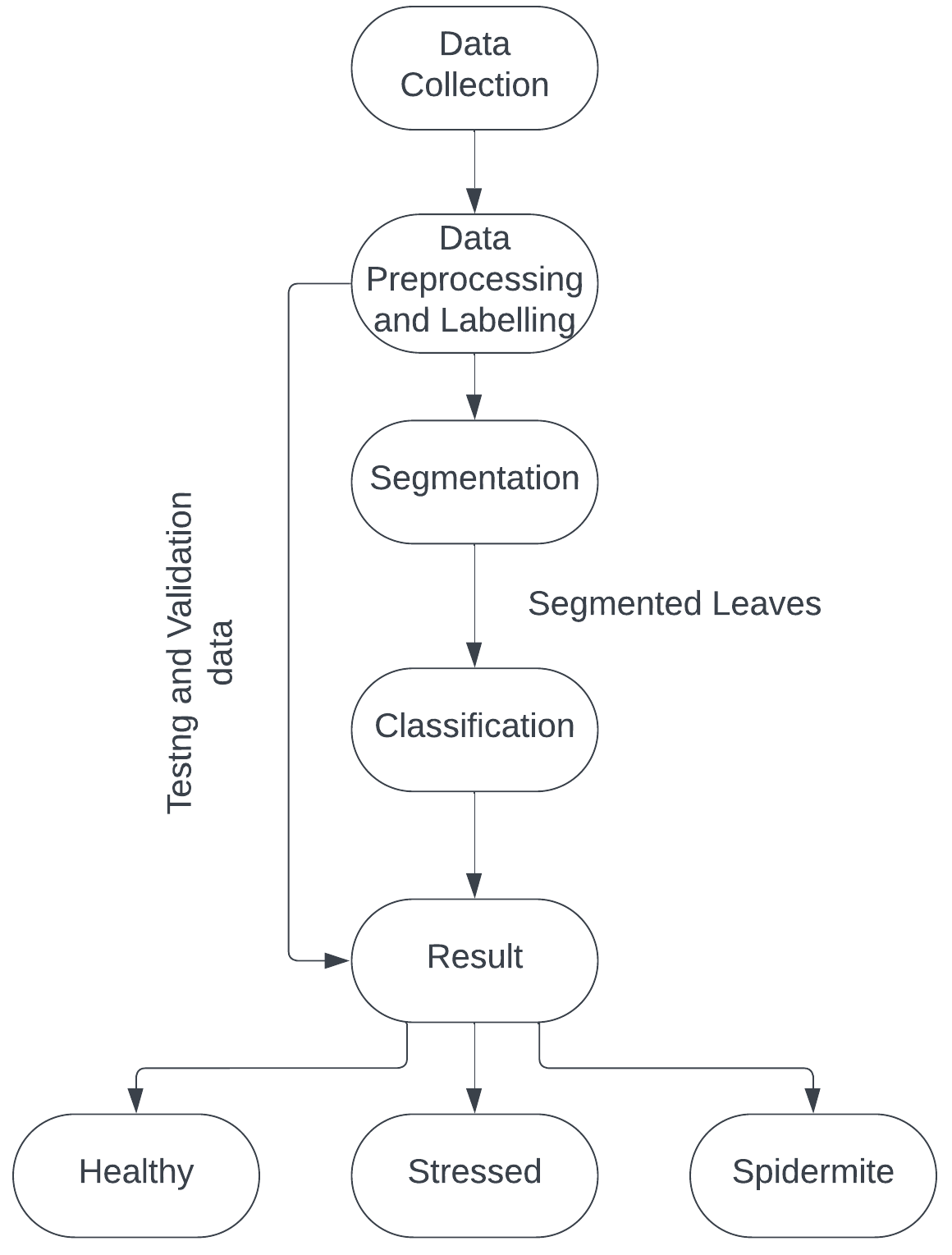} 
        \caption{Two-Stage Model}
    \end{subfigure}
    \caption{Comparison of the Single-Stage Model (a) and the Two-Stage Model (b).}
    \label{fig:workflow1}
\end{figure*}

Two models are investigated to streamline early plant disease detection while addressing the challenge of partially labelled data: a single-stage segmentation model (Figure \ref{fig:workflow1}a) utilising the "Occluded dataset" and a two-staged model (Figure \ref{fig:workflow1}b), separating the segmentation and classification processes for modularity. These two models will be compared to find the most optimal final model.

% \begin{figure*}[t]
%     \centering
%     \includegraphics[width=0.8\textwidth]{images/Blank diagram (1).png}
%     \caption{Visualisation of the Training Workflow (left) and the Proposed Model (right) }
%     \label{fig:workflow1}
% \end{figure*}

\subsection{Two-stage Model}
%jche774
The two-stage model comprises a segmentation model and a classification model; the segmentation model takes an image of a plant and separates the individual leaves from the background regardless of their conditions, and the classification model then takes each leaf image and assigns it a label based on the leaf condition, such as "Healthy", "Stressed" or "Spidermite". Combined, the two models act like a standard single-stage model, which segments healthy and diseased leaves from an image input but can be trained on a partially labelled dataset.

\begin{figure}[!htb]
    \centering
    \includegraphics[width=\columnwidth]{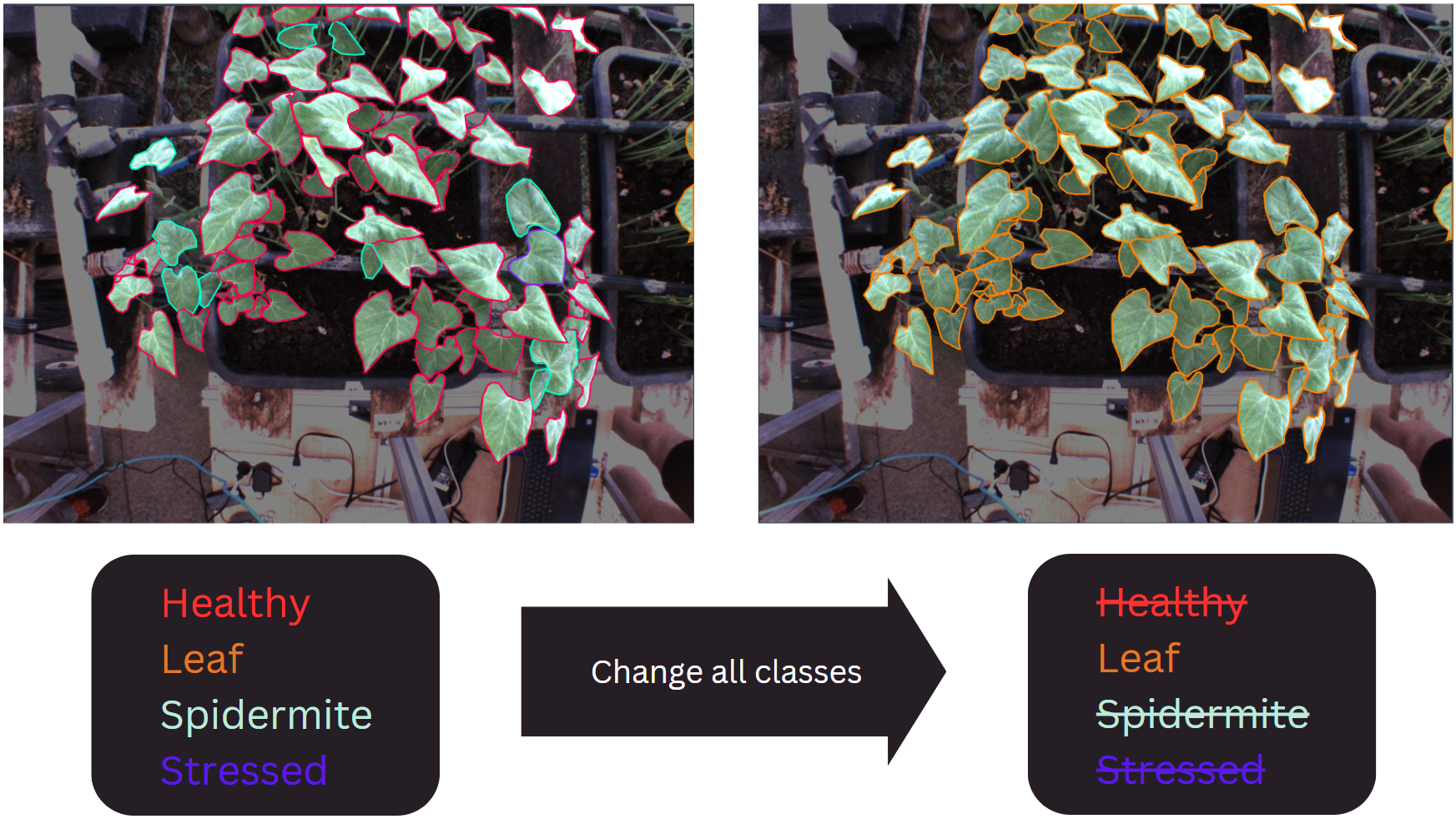}
    \caption{Dataset conversion for segmentation model.}
    \label{fig:segmentation_model_conversion}
\end{figure}

Two models are trained separately and require making some adjustments to the dataset. The segmentation model should exclusively identify leaves, so all considered classes in the training dataset must be consolidated into a general 'leaf' class as shown in Figure \ref{fig:segmentation_model_conversion}.  

\begin{figure}[!htb]
    \centering
    \includegraphics[width=\columnwidth]{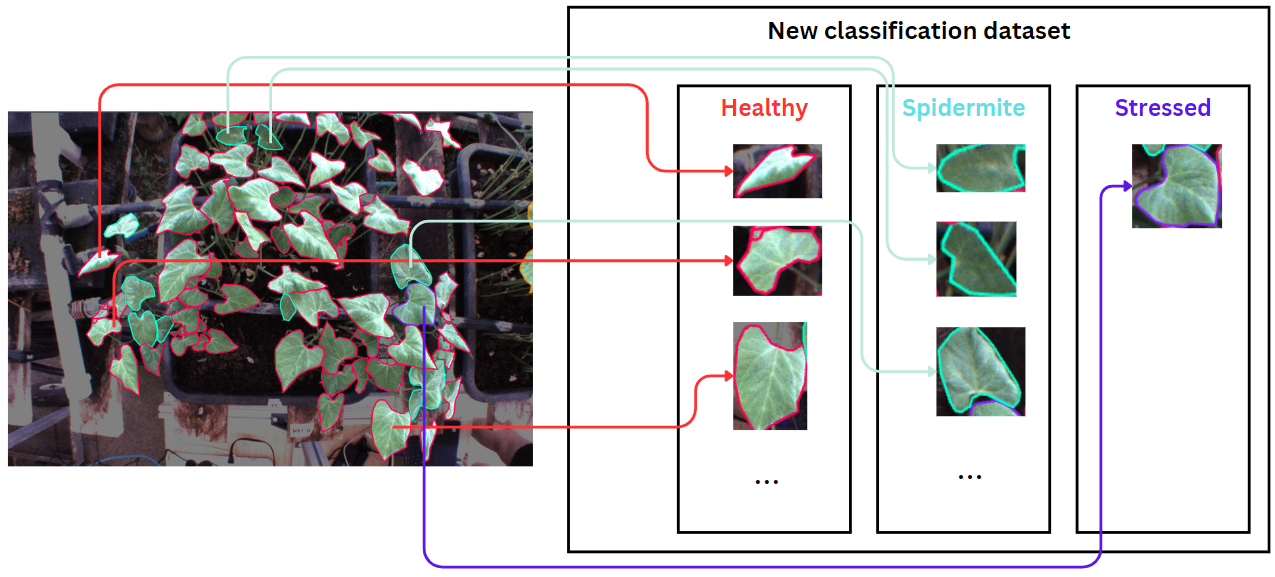}
    \caption{Dataset conversion for classification model.}
    \label{fig:classify_dataset_conversion}
\end{figure}

Conversely, the classification model aims to determine the health condition of each leaf and should only be trained on cropped and labelled leaf segments from the original dataset, as depicted in Figure \ref{fig:classify_dataset_conversion}

% \begin{figure*}[t]
%     \centering
%     \includegraphics[width=0.9\textwidth]{images/two_stage_model.png}
%     \caption{Architecture of designed two-stage model}
%     \label{fig:two_stage_model_architecture}
% \end{figure*}
The interference workflow of the two-stage model is shown in Figure \ref{fig:workflow1} (b). It receives an input image, then segments the image into 'leaf' segments and classifies their health conditions. After processing all segments, it delivers a final output, presenting the detected leaf instances and their respective health condition labels.

\subsection{Single-stage Model}
Another solution for the missing labels is to employ occlusions to conceal these unlabelled instances, rendering them black and easily distinguishable from the background; this approach enables training on a single-stage segmentation model to effectively acquire knowledge about the characteristics of the leaves without any interference from the unlabelled instances.

\begin{figure}[!hbt]
    \centering
    \includegraphics[width=\columnwidth]{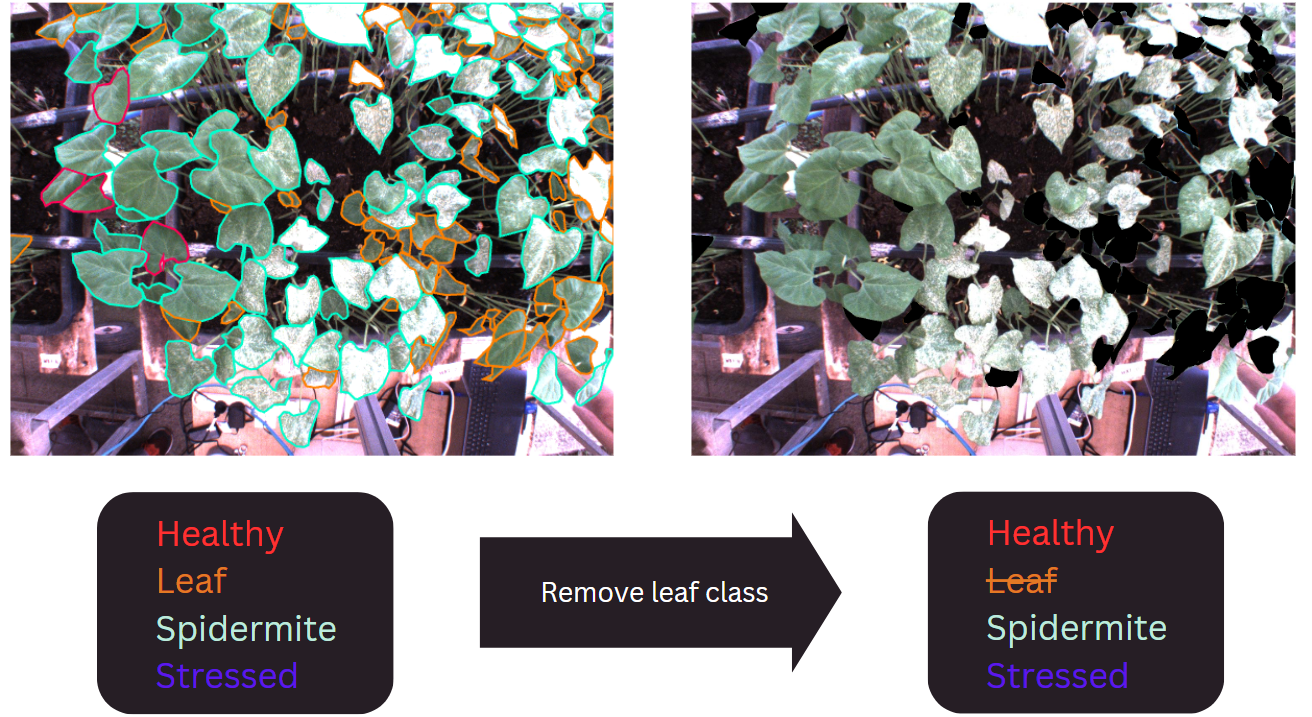}
    \caption{Removal of leaf class to obtain a fully labelled dataset.}
    \label{fig:leaf_class_removal}
\end{figure}

As shown in Figure \ref{fig:leaf_class_removal}, the leaf instances without ground truth in the original dataset are highlighted in orange. These instances are occluded with black pixels to generate a new dataset, and their labels will be removed.

\subsection{Segmentation}
The framework YOLOv8 was chosen to train the segmentation component of the two-stage model due to its fast inference time and comparable accuracy. In the context of the two-stage model, inference speed becomes a pivotal factor due to the added overhead associated with forwarding each leaf instance to the classification component. YOLOv8 is a relatively complex segmentation model that has a large number of weight parameters. 

Training these parameters from scratch is very time-consuming and inefficient. Using transfer learning to initialize the model weights based on pre-trained features extracted from RGB channels is more efficient; modifying the input CNN layer of the YOLOv8 model is necessary to segment images with 4 channels of RGBN. 

\subsection{Channel Fusing Techniques}
Training on RGBN dataset requires modifications on the input layer of the YOLOv8 model. Since the original YOLOv8 model only accepts inputs with 3 channels in RGB, the input channel parameter of the CNN layer is changed from 3 to 4; this way, the same architecture and weights of the YOLOv8 model can be used for the subsequent layers. Due to the demonstrated increase in performance through transfer learning, observing similar enhancements in training the RGBN model would be highly advantageous. The unavailability of publicly accessible RGBN datasets also implies that there are no pre-trained RGBN models readily accessible for weight transfer, in contrast to the common practice of transferring weights for RGB models. Therefore, attempts were made to transfer partial weights from a trained RGB model to the new models. Different partial weight-transferring approaches have been compared, which can be grouped by two-channel fusing techniques: substitution and addition.

\subsubsection{Substitution technique}
The "substitution" method replaces one of the RGB channels with the NIR channel in the dataset and trains the model as a usual RGB model with the new data; this method benefits from transfer learning, but it also sacrifices the information of the replaced channel, which may be important for the object of interest. Moreover, the pre-trained weights for the replaced channel may not be suitable for the new NIR channel.

\subsubsection{Addition Technique}
The "addition" method involves adding the additional NIR channel along with the RGB channels as an input to the model; this requires adding the NIR pixel value to the dataset and changing the input layer of the machine learning model.

The shape of the weight tensor for the input layer of the segmentation model (CNN) is determined by the number of filters, input channels, and the size of the filters. For example, if there are k filters, each with size $w \cdot h$, and the input has $c$ channels, then the weight tensor has the shape $k \cdot c \cdot w \cdot h$. Modifying the input layer to include an additional NIR channel will increase the value of $c$ by 1; this changes the shape of the weight tensor, preventing direct transfer learning. It remains feasible to leverage the pre-existing weights from the pre-trained model by transferring the pre-trained RGB weights to the new model while simultaneously introducing a new NIR channel without any pre-trained weights.

\subsection{Classification}
The current state-of-the-art classification algorithms in early plant disease detection are typically those of CNNS, in particular, that of VGG16 and ResNet \cite{Alatawi2022,Shah2022,saeed2021multi}. The labelled data was cropped into individual leaf-bound boxes of 256x256 pixels to train the classification models. The leaves were labelled into three classes: "Healthy", "Stressed" and "Spidermite". There were 89 images for "Healthy",  159 images for "Spidermite", and 84 images of "Stressed" leaves. Data was split into 80\% training, 10\% validation and 10\% testing. Four models were compared within this study: VGG16, ResNet15, ResNet50 and a simple sequential CNN network. These models were chosen as VGG16 and ResNet are two of the current state-of-the-art models in plant disease detection, and the simple sequential CNN network was used as a baseline for the study \cite{Alatawi2022,Shah2022}. A range of data configurations are explored. These configurations included:
\begin{itemize}
    \item RGB data using 3 channels.
    \item RGBN data using 4 channels.
    \item VGG16 and ResNet50 with transfer learning
    \item VGG16 and ResNet50 without transfer learning
\end{itemize}

However, since there is no publically available pre-trained VGG16 or ResNet50 with NIR data, the pre-trained models will only consider variations involving 3 channels. The sequential CNN model comprises 6 2D convolutional layers separated by max-pooling layers. Following this, the data is flattened and passed through a dense layer with Rectified Linear Unit (ReLU) activation, and subsequently, it flows into another dense layer with Softmax activation. All models share identical standard hyperparameters, featuring a consistent learning rate of 0.001 and a fixed batch size of 32 for 100 epochs.
\section{Results}
\subsection{Data Collection}
\begin{figure}[!htb]
    \centering
    \includegraphics[width=0.35\columnwidth]{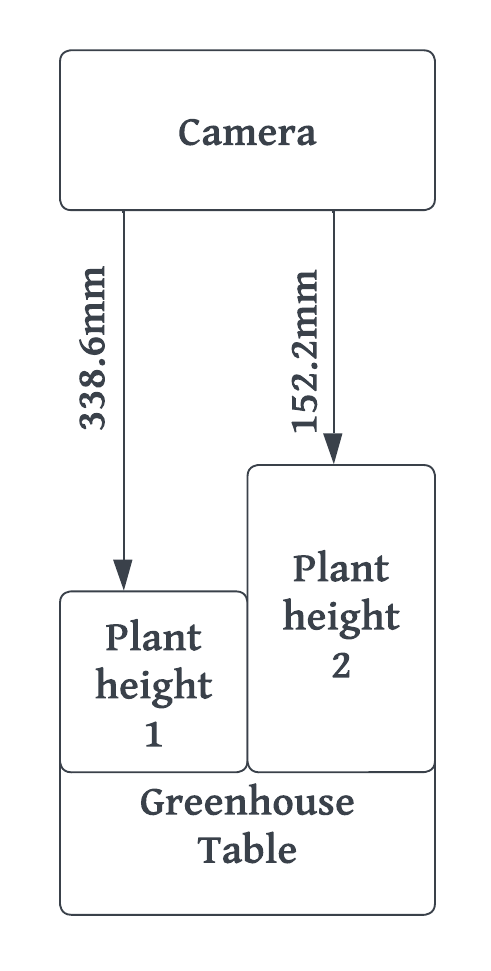}
    \caption{Visual representation of camera distance to plants.}
    \label{fig:Camera}
\end{figure}

In-field data collection mitigates dataset bias, ensuring an authentic representation of field-collected data; this is done using a JAI FS-1600D-10GE camera to collect data from labrador bean leaves from a BioForce greenhouse, as shown in Figure \ref{fig:greenhouse}. The camera has a resolution of 1440x1080 pixels and was configured with a 2.8mm focal distance. The camera-to-leaf distance ranged from 152.2mm to 338.6mm, while the NIR frequency fell within a 740nm to 1000nm spectrum. Figure \ref{fig:Camera} shows a visual representation of the camera setup. Through in-field data collection, 64 images were taken, 32 in RGB and 32 in NIR. Each image contains around 200 leaf instances to be labelled. The bean leaves were inoculated between 1 to 5 weeks before data collection, allowing the models to train for early disease detection, as the spider mite damage is yet to be visible to the naked eye.

One of the main challenges in training the segmentation and two-stage models is handling the dataset's missing labels. Accurately labelling the datasets poses a significant challenge, as leaves in the early stages of infection lack visible symptoms, necessitating external Polymerase Chain Reaction (PCR) testing for ground truth verification. Consequently, the dataset includes numerous leaves still undergoing ground truth testing, rendering them unsuitable for model training. Two basic approaches are ignoring unlabeled leaves in the image or training a segmentation model on the remaining labels. However, these methods are flawed because ignoring unlabeled leaves can result in misclassification, and creating a new 'leaf' class for unlabeled leaves would hinder the model's ability to differentiate between healthy and diseased leaves. Neither of these approaches fully addresses the problem.

\begin{figure}[h]
    \centering
    \includegraphics[width=\columnwidth]{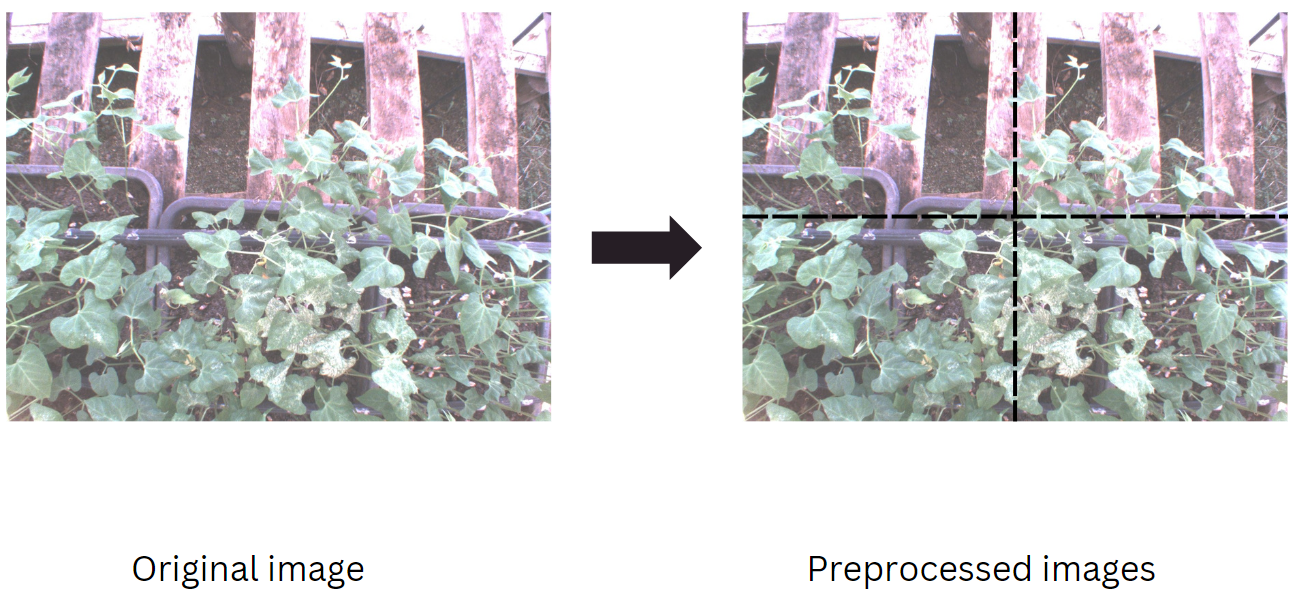}
    \caption{Image splitting on the dataset.}
    \label{fig:image_cutting}
\end{figure}

The original images in the dataset have a resolution of 1440x1080 pixels, which is too large for the training device's memory capacity. Therefore, all models discussed previously were trained on images resized to 480x360 pixels. However, this may affect the segmentation quality of small or distant objects. The limitation of segmentation size is resolved by generating a grid dataset, achieved by dividing each image into four equal parts, as shown in Figure \ref{fig:image_cutting}; this way, the models can learn more details from the original images with larger image sizes.

\subsection{Segmentation Component of Two-Stage Model}
A pre-trained YOLOv8 model trained on RGB segmentation through MS COCO dataset for 500 epochs achieves a mAP of 50\% by 50 epochs on the collected dataset. The identical model, configured with the same hyperparameters but without pre-trained weights, only achieved a mAP of 40.8\% after 483 epochs, with no further improvement observed during the last 50 epochs.

\subsubsection{Results of Channel Fusing Techniques}
Table \ref{tab:transfer_learning_substitution} shows the performance of different channel substitution methods to the NIR channel. A model trained with dataset channels of NGB means that the red channel of the images in the dataset is replaced by the NIR channel. After training with 100 epochs, the model without channel substitutions in the dataset performs the best with an mAP of 53.2\%.
\begin{table}[h]
    \begin{center}
        \caption{Channel substitution dataset on segmentation model.}
        \label{tab:transfer_learning_substitution}
        \begin{tabular}{|l|l|l|}
        \hline
        Dataset channels & Epochs & mAP50-95 \\ \hline
        RGB              & 100    & 53.2\%   \\ \hline
        NGB              & 100    & 51.1\%   \\ \hline
        RGN              & 100    & 48.7\%   \\ \hline
        \end{tabular}
    \end{center}
\end{table}
The following Table.\ref{tab:transfer_learning_segmentation} illustrates how different values of transferred weights affect the performance of the additional channel. Transferred weights indicate the degree to which the weights of the RGBN channel of the new CNN layers are derived from the pre-trained weights of the RGB model. A value of 'x' signifies that the weights on this channel are not pre-trained and are randomly initialized. For instance, RGBR in the table implies that the RGBN model uses the pre-trained weights of the R, G and B channels from the RGB model for the corresponding input CNN layer, and the fourth NIR channel inherits the weights from the red channel of the pre-trained RGB model.
\begin{table}[!htb]
    \caption{Transfer learning on segmentation model.}
    \label{tab:transfer_learning_segmentation}
    \centering
    \begin{tabular}{|l|l|l|}
        \hline
    Transferred weights in CNN & Epochs & mAP50-95                                                                  \\ \hline
    xxxx                & 100    & 48.5\%\footnotemark{}                   \\ \hline
    RGBx                & 100    & 52.2\%\footnotemark[\value{footnote}]\\ \hline
    RGBR                & 100    & 53.0\%                                                                    \\ \hline
    RGBG                & 100    & 51.2\%                                                                    \\ \hline
    RGBB                & 100    & 52.7\%                                                                    \\ \hline
    \end{tabular}
\end{table}
\footnotetext{Computed from an average of 3 training attempts since some weights are randomly initialised.}

Table \ref{tab:transfer_learning_segmentation} shows that training the RGBN model with pre-trained weights on the RGB channels can train the model more efficiently, with an average increase of 3\% in mAP within 100 epochs. However, the models have no significant difference with different transferred weights of the additional channel.

As shown in Table \ref{tab:transfer_learning_substitution}, the model with additional NIR does not outperform the standard RGB segmentation model with the same configurations, having a best-case mAP of 51.1\% and  53.2\% respectively. Contrary to initial expectations, the NIR channel does not enhance the model's robustness on segmenting leaves.

\subsubsection{Grid dataset}

Due to graphical memory limitations, a segmentation model can be trained with an image size of 704x528 with the grid RGB dataset for 50 epochs. Even with fewer training epochs, the model outperformed the one trained on the original dataset, as demonstrated in Table \ref{tab: model_training_on_grid_dataset}. Table \ref{tab: model_training_on_grid_dataset} indicates that training the segmentation model on the grid dataset is advantageous because it enables the model to capture finer leaf details, enhancing its ability to distinguish between leaves and the background; this is currently the best-performing segmentation model and will be used as the segmentation component in the two-stage model.

\begin{table}[h]
    \centering
    \caption{Model training on grid dataset.}
    \label{tab: model_training_on_grid_dataset}
        \resizebox{\columnwidth}{!}{%
            \begin{tabular}{|l|l|l|l|l|}
            \hline
            Dataset  & Image size & Epochs & mAP50-95 & Inference time(ms) \\ \hline
            Original & 480x360        & 100    & 53.2\%   & 5.8                \\ \hline
            Grid     & 704x528        & 50     & 63.0\%   & 23.5               \\ \hline
            \end{tabular}%
    }
\end{table}

\subsection{Classification Component of Two-Stage Model}
 The outcomes of training various models, including VGG16 with and without transfer learning, ResNet50 with and without transfer learning, ResNet15, and the sequential CNN model using the collected labrador bean leaf dataset, are summarised in Table \ref{tab:classification_result}.
\begin{table*}[]
\centering
\caption{Results of VGG16, ResNet50, ResNet15 and Sequential CNN model. Each training model was trained on the Labrador Beans Dataset with three classes, using the following hyperparameters: a learning rate of 0.001, 100 epochs, and a batch size of 32.}
\label{tab:classification_result}
\centering 
\resizebox{0.65\textwidth}{!}{%
\begin{tabular}{|l|c|c|c|c|c|c|}
\hline
\textbf{Model} & \textbf{\begin{tabular}[c]{@{}c@{}}Pre-trained? \\ (Y/N)\end{tabular}} & \textbf{RGB} & \textbf{NIR} & \textbf{\begin{tabular}[c]{@{}c@{}}\# of\\ channels\end{tabular}} & \textbf{\begin{tabular}[c]{@{}c@{}}Validation\\ Accuracy\end{tabular}} & \textbf{\begin{tabular}[c]{@{}c@{}}Validation\\ Loss\end{tabular}} \\ \hline
\textbf{VGG16} & N & Y & Y & 4 & 25\% & 1.1161 \\ \hline
\textbf{VGG16} & N & Y & N & 3 & 40.62\% & 4.1589 \\ \hline
\textbf{VGG16} & Y & Y & N & 3 & 87.5\% & 0.3476 \\ \hline
\textbf{Resnet50} & N & Y & Y & 4 & 25\% & 1.137  \\ \hline
\textbf{Resnet50} & N & Y & N & 3 & 25\% & 1.136 \\ \hline
\textbf{Resnet50} & Y & Y & N & 3 & 25\% & 1.1716 \\ \hline
\textbf{Resnet15} & N & Y & Y & 4 & 62.5\% & 0.7414 \\ \hline
\textbf{Resnet15} & N & Y & N & 3 & 56.25\% & 0.8752 \\ \hline
\textbf{Sequential} & N & Y & Y & 4 & 90.62\% & 0.2365 \\ \hline
\textbf{Sequential} & N & Y & N & 3 & 84.38\% & 0.5029 \\ \hline
\end{tabular}%
}
\end{table*}

\begin{figure}[htbp]
    \centering
    \begin{subfigure}[t]{0.48\columnwidth}
        \includegraphics[width=40mm]{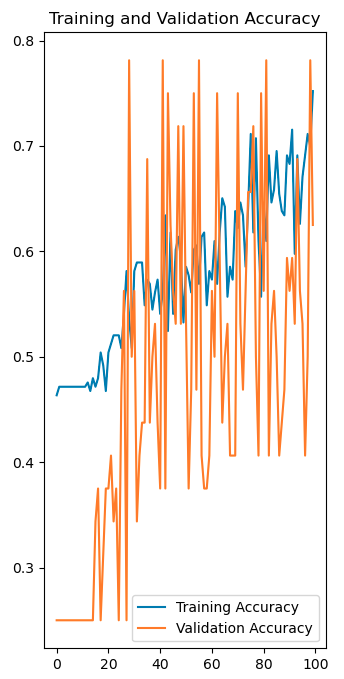}
        \caption{ResNet15 with 4 Channels.}
        \label{fig:resnet15rgbn}
    \end{subfigure}
    \begin{subfigure}[t]{0.48\columnwidth}
            \includegraphics[width=40mm]{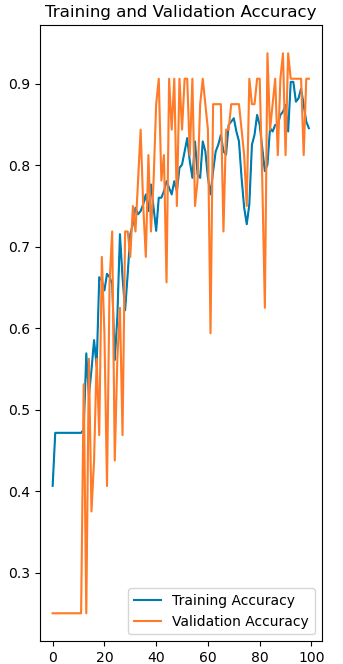}
            \caption{sequential CNN model with 4 Channels.}
            \label{fig:sequentialrgbn}
    \end{subfigure}
    \begin{subfigure}[t]{0.48\columnwidth}
        \includegraphics[width=40mm]{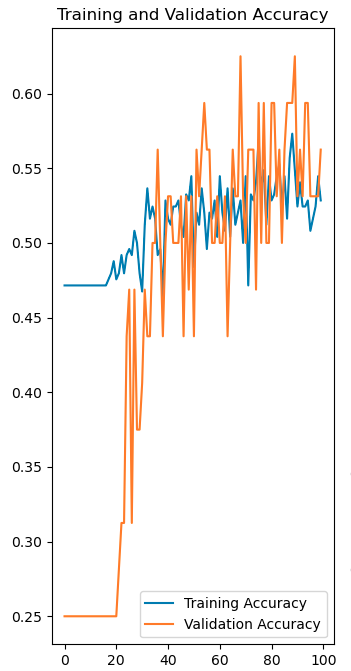}
        \caption{ResNet15 with 3 Channels.}
        \label{fig:resnet15rgb}
    \end{subfigure}
    \begin{subfigure}[t]{0.48\columnwidth}
        \includegraphics[width=40mm]{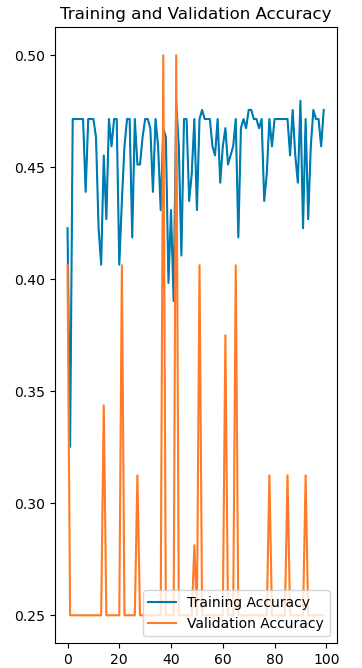}
        \caption{ResNet50 with 3 Channels.}
        \label{fig:resnet50rgb}
    \end{subfigure}
    \caption{Validation training on different classification models.}
\end{figure}

Through Table \ref{tab:classification_result}, it can be seen that the sequential CNN model with RGBN performs the best, achieving a 90.62\% validation accuracy at 100 epochs, followed by pre-trained VGG16 with 87.5\% validation accuracy using only RGB data. Notably, pre-trained ResNet50 models underperformed with a mere 25\% validation accuracy. Non-pretrained VGG16 with RGBN data and ResNet15 with RGB data also scored 25\% validation accuracy. However, ResNet15 improved to 62.5\% validation accuracy with 4-channel NIR data. Through Figures \ref{fig:resnet15rgbn} and \ref{fig:sequentialrgbn}, it can be seen that the ResNet15 with RGBN and sequential CNN RGBN both have yet to plateau, meaning that they have the potential for further improvement with extended training if not for hardware constraints.

Considering that the sequential CNN model has the highest validation accuracy, it will be the model utilised for the classification component of the two-stage model.

%\begin{figure}[!h]
%    \centering
%    \includegraphics[width=6cm, height=6cm]{images/resnet15rgbn.png}
%    \caption{Validation training of ResNet15 with 4 Channels.}
%    \label{fig:resnet15rgbn}
%\end{figure}
%\begin{figure}[!h]
%    \centering
%    \includegraphics[width=6cm, height=6cm]%{images/sequentialrgbn.png}
%    \caption{Validation training of the sequential CNN model with 4 Channels.}
%    \label{fig:sequentialrgbn}
%\end{figure}

%\begin{figure}[!h]
 %   \centering
 %   \includegraphics[width=5cm, height=4cm]{images/resnet15rgb.png}
%    \caption{Validation training of ResNet15 with 3 Channels}
%    \label{fig:resnet15rgb}
%\end{figure}
%\begin{figure}[!h]
%    \centering
%  \includegraphics[width=5cm, height=4cm]{images/rednet50rgb.png}
 %   \caption{Validation training of ResNet50 with 3 Channels}
%    \label{fig:resnet50pretrained}
%\end{figure}

\subsection{Single stage model performance}
The training results of the single-stage model on the occluded dataset are presented in Table \ref{tab:one_stage_occluded_dataset_evalution}, which includes three models: one trained on an RGB occluded dataset, another on an RGBN dataset with weight transfer from the red channel of a pre-trained model to the NIR channel (RGBR), and a third on the RGBN dataset with a randomly initialised fourth channel (RGBx). The training results across different models all achieved similar performances, as they all reached a saturation point with no further improvements beyond 32 epochs, leading to early termination.

\subsection{Evaluation on occluded dataset}
The single-stage and two-staged models are compared by utilising an occluded dataset. The comparison needs to be revised as both single-stage and two-staged models may be influenced by their individual preparation methods. However, the results are still comparable through careful analysis. The single-stage model is trained using the occluded dataset, enabling it to learn how to handle occlusions within the validation dataset, including occluded leaves.

The two-stage model approach is also evaluated on the same occluded dataset. The segmentation component is the model trained with the RGB grid dataset with an mAP of 63.0\%. The classification component is the sequential CNN model trained with the RGBN dataset, which achieved 90.62\%  validation accuracy. Experiments are also carried out on replacing the segmentation component with the single-stage model trained with different transferred weights. The segmentation component achieved the highest mAP of 23.6\% out of all other approaches.

Both the single-stage and two-stage approaches have varying advantages over the other. The single-stage model benefits from directly training from the occluded dataset, meaning it would be less likely to miscategorise black pixels as leaves. The two-stage model benefits from being tested on non-occluded leaves. However, the first-stage segmentation model may detect slightly differently than the ground truth labels on leaf segments and pass the leaves to the second-stage classification model with slight deviation.

The single-stage model gains an advantage by training directly on the occlusion dataset, allowing it to learn how to handle occluded leaves. This knowledge enables the model to deal effectively with occlusions in the validation dataset, including occluded leaves. The two-stage model has yet to be trained on the occlusion dataset, providing an advantage to the single-stage model. 

The classification component of the two-stage model faces a challenge as it was initially trained with bounding box data but now receives leaf masks as inputs when integrated into the segmentation component pipeline, causing a longer processing time.

Different from the two-stage model with a specialised classifier, the single-staged model cannot classify leaves effectively due to a lack of generalisation. 

\begin{table*}[t]
    \centering
    \caption{Single stage models performance on unseen occluded dataset with 1024x768 images.}
    \label{tab:one_stage_occluded_dataset_evalution}
    \resizebox{0.8\textwidth}{!}{%
        \begin{tabular}{|l|l|l|l|l|l|l|}
        \hline
        \textbf{Model} & \textbf{Epoch} & \textbf{\textbf{mAP (all)}} & \textbf{mAP (healthy)} & \textbf{mAP (spidermite)} & \textbf{mAP (stressed)} & \textbf{Inference time (ms)} \\ \hline
        RGB & 100 (32) & 17.8\% & 23\% & 30.5\% & 0.00\% & 90.4 \\ \hline
        RGBR & 100 (32) & 20\% & 27.6\% & 26.9\% & 5.69\% & 43.2 \\ \hline
        RGBx & 100 (32) & 18.55\% & 18.8\% & 36.4\% & 0.545\% & 68.9 \\ \hline
        \end{tabular}%
        }
\end{table*}

\begin{table*}[t]
\caption{Two-stage models performance on unseen occluded dataset.}
\label{tab:two_stage_occluded_dataset_evalution}
\centering % Center the table on the page
\resizebox{0.8\textwidth}{!}{%
\begin{tabular}{|l|c|c|c|c|c|c|}
\hline
\multicolumn{1}{|c|}{\textbf{\begin{tabular}[c]{@{}c@{}}Segmentation \\ model\end{tabular}}} & \textbf{\begin{tabular}[c]{@{}c@{}}Classification \\ model\end{tabular}} & \textbf{\begin{tabular}[c]{@{}c@{}}mAP \\ (all)\end{tabular}} & \textbf{mAP} & \textbf{\begin{tabular}[c]{@{}c@{}}mAP \\ (spidermite)\end{tabular}} & \textbf{\begin{tabular}[c]{@{}c@{}}mAP \\ (stressed)\end{tabular}} & \textbf{\begin{tabular}[c]{@{}c@{}}Inference \\ time (ms)\end{tabular}} \\ \hline
RGB single-stage model & RGBN model & 15.6\% & 19.6\% & 23.2\% & 4.07\% & 140 \\ \hline
RGBx single-stage model & RGBN model & 16\% & 23.4\% & 16.3\% & 8.29\% & 430 \\ \hline
RGBR single-stage model & RGBN model & 16.5\% & 25\% & 19.3\% & 5.27\% & 240 \\ \hline
RGB grid two-stage model & RGBN model & 23.6\% & 40\% & 30.7\% & 0.00\% & 290 \\ \hline
\end{tabular}%
}
\end{table*}

\section{Discussion}
\subsection{Segmentation}
Table \ref{tab:transfer_learning_substitution} shows that the substitution channel fusing technique with transfer weights is not always reliable, in which either replacing the red channel or green channel did not produce comparable results as the RGB dataset; this proves that the red and blue channels continue to offer valuable information for distinguishing leaves, contributing to the model's accuracy. One plausible explanation is that the availability of a large RGB dataset has provided an incomparable advantage when training the RGB model, surpassing the potential benefits that a NIR channel might offer in discovering additional features. The substitution channel model results suggest that transferring input layer weights from any RGB channel in the pre-trained model does not enhance NIR channel training; this is supported by Table \ref{tab:transfer_learning_segmentation}, where models using RGBN data perform similarly to those without NIR channel transfer learning, indicating that existing RGB channels are unsuitable for pre-training the NIR channel. Nonetheless, partial transfer learning's success is evident compared to models without input layer pre-training, as shown in Table \ref{tab:transfer_learning_segmentation}.
\subsection{Classification}
During testing, it became clear that incorporating RGBN data outperforms using only RGB data, as shown in Table \ref{tab:classification_result}. The improvement is likely attributed to the additional information provided by the NIR channel, enabling the model to discern finer distinctions \cite{Liu2020,Nieuwenhuizen2020}. Interestingly, the sequential CNN model performs best among the three proposed models; this advantage likely results from various factors, such as the dataset size, small leaf image size, low training epochs, and differences in model complexity between VGG16 or ResNet50 and ResNet15 or sequential CNN. As evidenced by the performance gap between pre-trained and non-pre-trained VGG16 models, the reduced validation accuracy without pre-training can be attributed to a densely connected network requiring more training epochs when initiated from scratch.
Conversely, this phenomenon is less pronounced with ResNet50, likely due to its larger trainable parameters, necessitating more data and training time to prevent overfitting, excessive data analysis and proper training \cite{Brigato2021}. To further exemplify this hypothesis, ResNet15 was implemented due to it containing relatively less trainable parameters than that of ResNet50. Figure \ref{fig:resnet15rgb} and \ref{fig:resnet50rgb} show the training progress of 3 channel ResNet15 compared to that of the sequential CNN model and ResNet50.

%\begin{figure}[!h]
%    \centering
%    \includegraphics[width=5cm, height=4cm]{images/resnet15rgb.png}
%    \caption{Validation training of ResNet15 with 3 Channels}
%    \label{fig:resnet15rgb}
%\end{figure}
%\begin{figure}[!h]
%    \centering
%    \includegraphics[width=5cm, height=4cm]{images/rednet50rgb.png}
%    \caption{Validation training of ResNet50 with 3 Channels}
 %   \label{fig:resnet50pretrained}
%\end{figure}

These trends show that ResNet50 is stagnant within local minima, likely due to its many trainable parameters. While ResNet15 is still gradually learning, its validation accuracy remains low even after 100 epochs. In contrast to these relatively complex models, the sequential CNN model steadily increases in validation accuracy, supporting the notion that more complex CNN models may struggle to learn when exposed to limited data. Analyzing the validation accuracies of models that did not stagnate at local optima, it becomes evident that incorporating RGBN data results in an average increase of 6.25\% in detection validation accuracy compared to using only RGB data; this strongly indicates that including NIR data provides valuable information, enabling the classification model to detect diseases at an earlier stage than when relying solely on RGB data \cite{rous,Nieuwenhuizen2020,Liu2020}.

\subsection{Model evaluation on occluded dataset}
The two-stage model, featuring the most accurate leaf segmentation and disease classification models as indicated in Table \ref{tab:two_stage_occluded_dataset_evalution}, outperforms the single-stage model presented in Table \ref{tab:one_stage_occluded_dataset_evalution} by achieving an improved mAP of 3.6\% The experiments replacing the leaf segmentation model with the single-stage model, as depicted in Table \ref{tab:two_stage_occluded_dataset_evalution}, have demonstrated that the leaf segmentation model improved its ability to distinguish leaves by training on additional leaves lacking ground truth information about their conditions. However, this comes with the sacrifice of speed as two-stage model relies on passing every segments to the classification component. From both tables difference in inference time per image of the two models is found to be magnitude, and the inference time of two-stage model can increase as the number of instances detected increases.

Nevertheless, this evaluation result requires careful interpretation due to the limitations of the occluded dataset, which only contains 5 training images and 2 validation images. It is expected that the model trained on the dataset cannot generalize well, and the validation on the dataset would not be comprehensive. 

\section{Conclusion and Future Works}
Early plant disease detection is vital amid increasing global food demand, with pests and diseases causing significant crop yield losses. This study explores methods like RGB and NIR channels, machine learning models, and specialized datasets for disease detection. NIR data shows promise in enhancing accuracy, while dataset challenges, like biases and variations, are acknowledged. A two-stage model for handling missing labels is proposed, with a sequential CNN model achieving a strong validation accuracy of 90.62\% with RGBN data. Future work should involve advanced hardware, larger datasets, and more complex models to address global food challenges.

In future, the studies should be expanded upon by using more advanced hardware to allow for more training epochs for classification. It would also be beneficial to collect more data to increase the size of the dataset to allow for the implementation of more complex models.

% collect more data
% occlude better 
% \section*{Acknowledgment}

% The preferred spelling of the word ``acknowledgment'' in America is without 
% an ``e'' after the ``g''. Avoid the stilted expression ``one of us (R. B. 
% G.) thanks $\ldots$''. Instead, try ``R. B. G. thanks$\ldots$''. Put sponsor 
% acknowledgments in the unnumbered footnote on the first page.
\section*{Acknowledgements}
    We extend our thanks to Bioforce Limited and their representative, Chris Thompson, for sponsoring this project and supplying resources to assist with the research.

\bibliographystyle{named}
\bibliography{references}
%\vspace{2pt}
\end{document}